\documentclass{article}

\usepackage[english]{babel}

\usepackage[a4paper,top=2cm,bottom=2cm,left=3cm,right=3cm,marginparwidth=1.75cm]{geometry}

\usepackage{amsmath}
\usepackage{graphicx}
\usepackage[colorlinks=true, allcolors=blue]{hyperref}
\usepackage[dvipsnames]{xcolor}

\usepackage{fancyhdr}
\fancypagestyle{firstpage}{
    \chead{Manuscript accepted to LocBigDataAI 2023} 
}

\title{Towards eXplainable AI for Mobility Data Science}
\author{Anahid Jalali, Anita Graser, Clemens Heistracher
\\ \small{anahid.jalali | anita.graser | clemens.heistracher@ait.ac.at}
\\  \small{AIT Austrian Institute of Technology GmbH}}
\date{}

\begin{document}
\maketitle
\thispagestyle{firstpage}

\begin{abstract}
This paper presents our ongoing work towards XAI for Mobility Data Science applications, focusing on explainable models that can learn from dense trajectory data, such as GPS tracks of vehicles and vessels using temporal graph neural networks (GNNs) and counterfactuals. We review the existing GeoXAI studies, argue the need for comprehensible explanations with human-centered approaches, and outline a research path toward XAI for Mobility Data Science.
\\ \\

\end{abstract}

\section{Introduction to GeoXAI}

XAI, or Explainable AI, develops Artificial Intelligence (AI) systems that can explain their decisions and actions. XAI thus promotes transparency and aims to enable trust in AI technologies~\cite{rai2020explainable}.
%
%
While traditional interpretable machine learning (ML) approaches (such as Gaussian Mixture Models~\cite{graser2020m3}, K-Nearest Neighbors~\cite{cheng2018spatiotemporal}, and decision trees~\cite{wylie2019geospatial}) have been widely used to model geospatial (and spatiotemporal) phenomena and corresponding data, the increasing size and complexity of spatiotemporal data have raised the need for complex methods to model such data. Therefore, recent studies focused on using black-box models, often in the form of deep learning models~\cite{graser_deep_2023, hsu2023explainable, feng2018deepmove, gao2019predicting, li2020predicting, bach2015pixel}. 

With this rise of Geospatial AI (GeoAI), there is a growing need for explainability, particularly for GeoAI applications where decisions can have \textit{significant social and environmental implications}~\cite{degas2022survey, xing2023challenges, cheng2021method}. 
However,  XAI research and development tends towards computer vision, natural language processing, and applications involving tabular data (such as healthcare and finance)~\cite{theissler2022explainable} 
and few studies have deployed XAI approaches for GeoAI (GeoXAI)~\cite{hsu2023explainable, xing2023challenges}.

\section{GeoXAI for Mobility}

To comprehend the challenges of integrating XAI approaches in GeoAI (for GIScience in general and Mobility Data Science~\cite{mokbel_mobility_2022} in particular) it is crucial to understand both the concepts and methods of XAI as well as the structure of and information contained in spatiotemporal data. 
%
%
%
Some of the challenges that must be addressed to realize GeoXAI include~\cite{xing2023challenges}: 

\begin{enumerate}
    \item Selection of appropriate reference data and models,
    \item Limitations of using gradients as explanations, 
    \item Difficulties in handling geographic scale,
    \item Inability to seamlessly integrate topology and geometry into the explanatory process of existing XAI tools (such as LIME and SHAP),
    \item Challenges in visualizing geography within XAI,
    \item Untapped potential of incorporating geospatial semantics and ontologies, and
    \item Lack of consideration for social and ethical aspects within XAI.
\end{enumerate}

To overcome these challenges, our Mobility GeoXAI research plan includes developing suitable models and XAI methods. 

\subsection{Our Mobility GeoXAI approach}

A promising deep learning \textbf{model} for mobility (addressing challenge 1) is temporal graph neural networks (TGNNs). 
Recent studies show that TGNNs capture temporal dependencies and relational information in mobility data, enabling, for example, accurate modeling and prediction of vessel trajectories~\cite{altan2022discovering, derrow2021eta, lippert2022learning}. To handle scale (challenge 3), we plan to investigate TGNN aggregation options to represent spatial/spatiotemporal scale.

Our working hypothesis for developing  \textbf{explanations} (challenge 2) is that examples are better at explaining model decisions (particularly for end users) than gradients/heatmaps (which are more targeted towards ML developers~\cite{hsu2023explainable}). Counterfactual examples involve generating alternative trajectories or scenarios to understand the impact of changes in the input data, for exmaple, by manipulating trajectory  attributes and geometry (challenge 4).  The Counterfactual analysis allows us to observe how the model's predictions would differ under alternative conditions. These examples answer why a specific prediction was made and what needs to be changed for a different prediction.
For instance, suppose we have an AIS trajectory data set and a TGNN model trained to predict vessel destinations. We can generate counterfactual trajectories by modifying specific attributes, such as altering the speed or route of a vessel. We can understand how specific factors influence the model's predictions by comparing the predicted destinations under different scenarios. This insight can identify critical features contributing to specific outcomes, enabling stakeholders to make informed decisions based on the model's explanations.
We believe that combining TGNNs and example-based explanations enhances the explainability of trajectory-based mobility models. 

Geospatial and mobility domain expert feedback will be engaged to develop and evaluate the required spatiotemporal \textbf{visualizations} for GeoXAI (challenge 5) and to integrate spatiotemporal semantics (challenge 6). And finally, privacy-preserving methods need to be adapted to ensure that GeoXAI accounts for the \textbf{privacy} protection necessary for mobility data (challenge 7).

\section{Related work}

Existing XAI research  mainly focused on computer vision tasks and tabular data, which are easier for non-domain experts to interpret than model decisions based on other data types, such as time series and geospatial data, which often require domain-specific knowledge~\cite{xing2023challenges, theissler2022explainable}. 
Nonetheless, GeoAI researchers have tried to enhance the interpretability and explainability of GeoAI and a good review of explanation types applicable to different GeoAI models is presented by Xing et al.~\cite{xing2023challenges}. 

However, most GeoXAI publications are still closely \textbf{related to computer vision} and therefore not readily transferable to mobility data. For example, Hsu and Li.~\cite{hsu2023explainable} generate saliency maps based on two model explanation methods: perturbation-based (manipulating input images) and gradient-based (visualizing model weights). 
Similarly, Li et al.~\cite{li2022extracting} utilized the Shapley Additive exPlanations (SHAP)~\cite{lundberg2017unified}, to compare the spatial effects extracted by XGBoost with traditional statistical approaches like the spatial lag model and multi-scale geographically weighted regression (MGWR). 
And, Xing and Sieber~\cite{xing2021integrating} employed SHAP 
to visualized the importance of feature maps at different stages of the convolution process. The authors noted the challenge of linking these explanations to semantic or geographic concepts, highlighting the need for deeper integration between XAI and GeoAI to understand the role of location and geographic attributes in modeling and decision-making processes.

\textbf{Mobility-specific XAI} is presented by Degas et al.~\cite{degas2022survey} reviewing XAI for Air Traffic Monitoring applications; and 
Cheng et al.~\cite{cheng2021method} proposed a method using Layerwise relevance Backpropagation (LRP)~\cite{bach2015pixel} to explain the significance of units within a Deep Neural Network when analyzing trajectory data for multiple estimated time of arrival (ETA) tasks. Other studies (such as~\cite{feng2018deepmove, gao2019predicting, li2020predicting}) use attention-based approaches to explain the decision of the models on trajectory data for applications, such as next location and final destination prediction. However, several studies argue that these attention-based approaches are not adequate to explain the decision made by the model for two reasons: first, they are not comprehensible by the end users~\cite{jain2019attention, wiegreffe2019attention}, and second, the provided attention maps are unreliable when manipulating the input data and can produce contradicting explanations~\cite{shi2021corpus}. 

Overall, these studies contribute to the ongoing exploration of XAI in the context of GeoAI, providing insights into the spatial effects captured by machine learning models, visualizing influential features, and emphasizing the importance of integrating XAI with geographic concepts and attributes. 

\section{Outlook}

Our ongoing work aims to apply the Mobility XAI concepts outlined above to a maritime mobility use case, specifically, detecting dark vessels (ships that illegally turn off their AIS signal). 
For this purpose, we use AIS trajectory data to train TGNNs. We explore the transformation of trajectories into graph structures using discretization techniques, for example, employing global grids, such as H3 (similar to ~\cite{wang2019machine}). To evaluate the performance of TGNNs on this dark vessel detection task, we furthermore have to address the need for expert input by incorporating a human-in-the-loop approach in our modeling pipeline. We plan to use counterfactual examples to enhance explainability, guide human experts, and incorporate their feedback into the ML pipeline.

\section{Acknowledgement} 
This work is mainly funded by the EU’s Horizon Europe research and innovation program under Grant No. 101070279 MobiSpaces, No. 101093051 EMERALDS, and No. 101021797 STARLIGHT.

\bibliographystyle{abbrv}
\bibliography{sample}

\end{document}